\documentclass[11pt]{article}
\usepackage[table]{xcolor}
\usepackage[preprint]{acl}

\usepackage{times}
\usepackage{graphicx}
\usepackage{subcaption}
\usepackage{latexsym}
\usepackage{setspace}
\usepackage{amsmath}
\usepackage{amsfonts}
\usepackage{placeins}
\usepackage{float}
\usepackage{booktabs} 
\usepackage{graphicx}
\usepackage{makecell}
\usepackage{marvosym}
\usepackage{authblk}
\usepackage{fontawesome}
\usepackage{enumitem}
\usepackage{setspace}
\definecolor{darkgreen}{rgb}{0.0, 0.5, 0.0}
\newcommand{\OurP}{RCoT }
\usepackage{amssymb}
\usepackage{pifont}
\usepackage[absolute,overlay]{textpos}
\usepackage{multirow}

\usepackage[T1]{fontenc}

\usepackage[utf8]{inputenc}

\usepackage{microtype}

\usepackage{inconsolata}

\usepackage{graphicx}

\title{UTMath: Math Evaluation with Unit Test \\via Reasoning-to-Coding Thoughts}

\author{
 \textbf{Bo Yang\textsuperscript{1}},
 \textbf{Qingping Yang\textsuperscript{2}},
 \textbf{Yingwei Ma\textsuperscript{2}},
 \textbf{Runtao Liu\textsuperscript{3}}
\\
 \textsuperscript{1}South China University of Technology\\
 \textsuperscript{2}ReasonMind\\
 \textsuperscript{3}Hong Kong University of Science and Technology
\\
 \small{
    \href{mailto:sdyangbo02@mail.scut.edu.cn}{sdyangbo02@mail.scut.edu.cn},
    \href{mailto:qingping95@gamil.com}{\{qingping95},
    \href{mailto:yingwei.ywma@gmail.com}{yingwei.ywma},
    \href{mailto:runtao219@gmail.com}{runtao219 \}@gmail.com}
 }
\\
\\
\href{https://utmathhomepage.github.io/}{https://utmathhomepage.github.io/}
}

\begin{document}
\maketitle
\begin{abstract}
The evaluation of mathematical reasoning capabilities is essential for advancing Artificial General Intelligence (AGI). 
While Large Language Models (LLMs) have shown impressive performance in solving mathematical problems, existing benchmarks such as GSM8K and MATH present limitations, including narrow problem definitions with specific numbers and reliance on predetermined rules that hinder accurate assessments of reasoning and generality. 
This paper introduces the UTMath Benchmark, a robust evaluation framework designed to assess LLMs through extensive unit tests, with a focus on both the accuracy and generality of model responses. It comprises 1,053 cutting-edge problems spanning nine mathematical domains, with an average of 68 test cases per problem. UTMath is highly challenging, with the best-performing model, o1-mini, solving only 32.57\% of the problems, followed by o1-preview at 27.16\%, and GPT-4o at 26.93\%.
Furthermore, we present the Reasoning-to-Coding of Thoughts (RCoT) approach, which encourages LLMs to engage in explicit reasoning prior to code generation, thereby facilitating the production of more sophisticated solutions and enhancing overall performance and efficiency. Furthermore, we also release the UTMath-Train training dataset (more than 70k samples), to support the community in further exploring mathematical reasoning.
Our benchmark can be accessed via the following link: \href{https://github.com/UTMathGroup/UTMath}{UTMath}

\end{abstract}

\section{Introduction}

The pursuit of AGI necessitates strong mathematical reasoning capabilities, making the evaluation of such abilities a crucial area of research~\cite{zhou2024your}. 
Recent advancements in LLMs have demonstrated remarkable proficiency in solving complex mathematical problems, achieving amazing performance on various datasets of Math Word Problems (MWPs), such as GSM8K~\cite{cobbe2021training}, MATH~\cite{hendrycks2021measuring}, TheoremQA~\cite{chen2023theoremqa}. 

\begin{figure}
    \centering
    \includegraphics[width=0.95\linewidth]{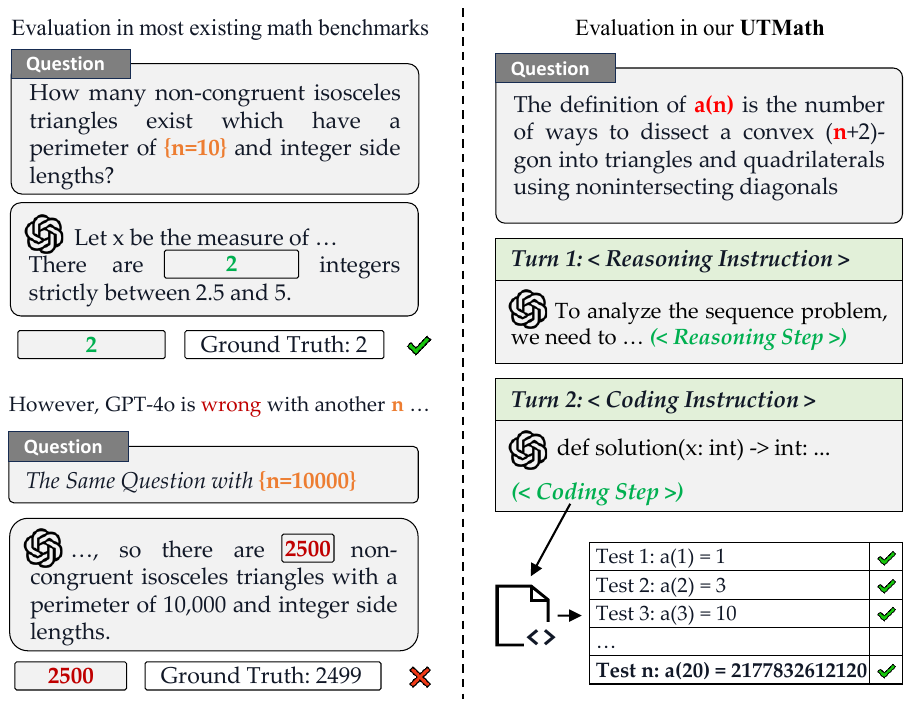}
    \caption{
    Comparison of UTMath with other benchmarks. On the left, GPT-4o easily solved one question but failed with a different numeric input. On the right, our benchmark is shown, where each problem includes multiple test cases, and a solution is correct only if all test cases are passed.
    We also propose a new prompting method RCoT in which the LLM first reasons through the problem and then generates code.}
    \label{fig:dataset_compare}
\end{figure}

However, classic benchmarks exhibit several limitations that impede the accurate and comprehensive assessment of these models' capabilities~\cite{ahn2024large}.
First, these benchmarks test models on narrowly defined problems with some specific numbers, which do not adequately assess generality to similar but varied scenarios as shown in Fig.~\ref{fig:dataset_compare}. 
Second, their evaluation relies on predetermined rules or the method of LLM-as-a-Judge(~\cite{dubois2024length, zheng2023judging}) that usually failed with capricious responses of LLMs. For example, an accurate answer need to be extracted to exactly match the fianl answer in the dataset GSM8K, TheoremQA, and MATH dataset.
While recent work has made great progress in developing new benchmarks, many of these approaches still fall short of addressing the fundamental limitations of earlier datasets. For instance, benchmarks like GSM-HARD~\cite{gao2023pal}, GSM-IC~\cite{shi2023large}, GSM-Plus~\cite{li2024gsm}, MetaMath~\cite{yu2023metamath} have extended the dataset of GSM8K or MATH with some perturbation such as substitution, reversing, distractor insertion. 
These efforts, while valuable, are characterized by limited coverage and high costs.
In this context, our work seeks to bridge these gaps by proposing a solid and robust benchmark that accurately evaluates the mathematical capabilities of LLMs.

Drawing on evaluation methods from software development, we propose the design of a comprehensive set of unit tests for mathematical problems to rigorously assess the reasoning processes of LLMs. If a solution successfully passes all unit tests within a class of problems, it suggests that the reasoning underlying the solution is more reliable and trustworthy.
Specifically, we introduce the \textbf{UTMath}, a novel benchmark derived from the On-Line Encyclopedia of Integer Sequences (OEIS)~\cite{OEIS2024}. 
The benchmark consists of 1,053 cutting-edge problems spanning 9 mathematical domains, such as Number Theory and Geometry. Each problem is accompanied by more than 68 test cases that provide a set of inputs and their corresponding outputs. 

In terms of evaluation methodology, our benchmark requires models to derive a general solution for a class of problems, typically represented in the form of code.
Compared to solving a problem defined by specific numbers, developing such a general solution is substantially more challenging, requiring higher levels of intelligence and reasoning ability.
However, when we performed ``Program of Thoughts (PoT)''~\cite{chen2022program}, wherethe model is required to perform reasoning and coding in a single response, it consistently tends to produce simpler and straightforward solutions. We surmise that this tendency may be influenced by the distribution of coding data. 
To address this, we introduced the \textbf{``Reasoning-to-Coding of Thoughts (RCoT)''}, which requires the LLM to perform mathematical reasoning in the first turn without any coding instruction, then writing code based on the reasoning. 
Compared to PoT, RCoT shifts the code distribution towards mathematics in the first turn, prompting more reasoning steps. 
\begin{figure}
    \centering
    \includegraphics[width=0.8\linewidth]{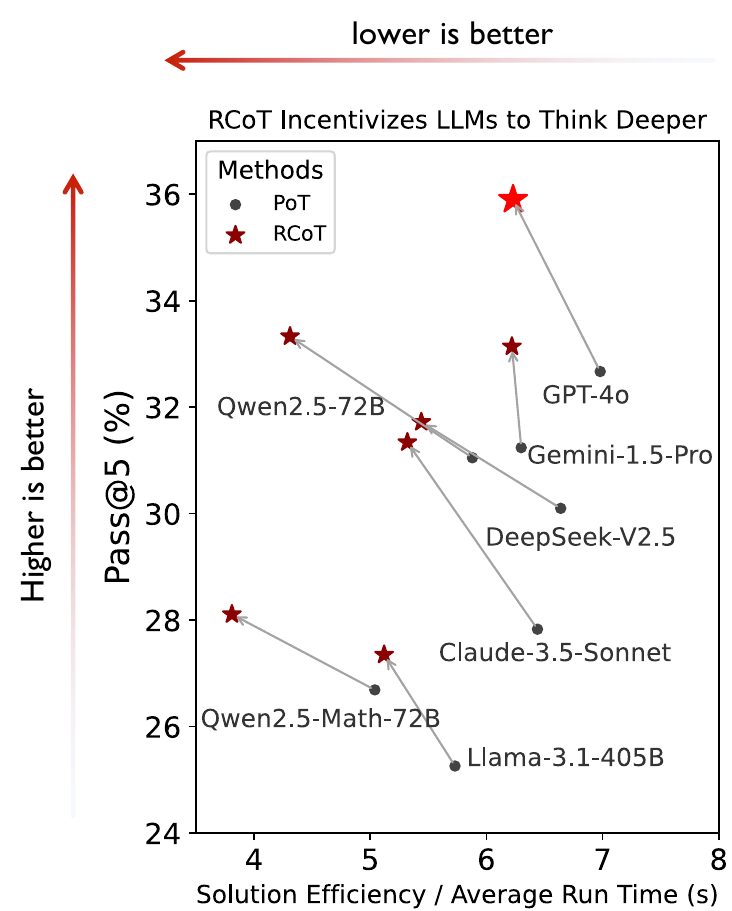}
    \caption{RCoT significantly improves the efficiency and effectiveness of the solution. It indicates that our \OurP proves to be more effective, suggesting that it encourages the model to reason critically and find more efficient solutions.}
    \label{fig:RCoT_results}
\end{figure}

UTMath is highly challenging, 

We conducted a comprehensive study with 8 LLMs. Some of our key findings are summarized as follows:
(1) with the best-performing model, o1-mini, solving only 32.57\% of the problems, followed by o1-preview at 27.16\%, and GPT-4o at 26.93\%, these results demonstrate the difficulty of UTMath. 
(2) Modern LLMs perform poorly in Graph Theory, Group Theory, Geometry and Topology (Fig.~\ref{tab:performance on different categories}). 
(3) With RCoT, all evaluated LLMs generated more efficient solutions, with most models achieving higher scores (Fig.~\ref{fig:RCoT_results}). 
(4) RCoT can significantly improve the pass@k performance of LLMs (\S~\ref{sec:exp_pass@k}). 
(5) The quality of reasoning significantly impacts the accuracy and efficiency of the model's final solution (\S~\ref{sec:exp_reasoning}). 
More interesting findings can be found in \S~\ref{sec:exp}. 
We hope our findings contribute to a deeper understanding of current reasoning ability of LLMs and the further development of models.

\section{Related Work}
\subsection{Benchmarks}

With the rapid development of LLMs, evaluating and exploring the intelligence and limitations of these models has emerged as an urgent issue to address ~\cite{chang2024survey}. Reasoning ability, as a crucial component of general intelligence, has garnered widespread attention since the advent of LLMs ~\cite{patel2021nlp, cobbe2021training, valmeekam2022large, perez2022discovering, gupta2022john, shakarian2023independent}. Mathematical reasoning, due to its complex mathematical characteristics and rigorous logical relationships, is considered an abstract and high-difficulty task, playing a pivotal role in demonstrating a model's reasoning capabilities.

To this end, researchers have proposed various benchmarks focused on mathematical reasoning. A natural and mainstream approach is to evaluate LLMs as humans would take math exams, using human exam questions to test their reasoning abilities, categorized by required knowledge levels. Examples include GSM8K at elementary school level, Math and GaokaoBench-Math~\cite{zhang2023evaluating} at high school level , College Math ~\cite{tang2024mathscale}, TheoremQA ~\cite{chen2023theoremqa}, ARB ~\cite{sawada2023arb} at university level, and OlympiadBench ~\cite{he2024olympiadbench}, AGIeval-Math ~\cite{zhong2023agieval} at competition level.

Besides, researchers have also introduced many others focused on evaluating various aspects of LLMs like the robustness. These include GSM8K-based variants: GSM-8K-Adv ~\cite{anantheswaran2024investigating}, GSM-Hard ~\cite{gao2023pal}, GSM-Plus ~\cite{li2024gsm}, GSM-IC ~\cite{shi2023large}, and several independent benchmarks: RobustMath ~\cite{zhou2024mathattack}, MetaMathQA ~\cite{yu2023metamath}, PROBLEMATHIC ~\cite{anantheswaran2024investigating}, MATHCHECK ~\cite{zhou2024your}, as well as other benchmarks ~\cite{li2024crowdsourced, li2023you}.

The distinctions between our proposed benchmark and existing ones are as follows. 
(1) Multiple Case Validation. Instead of using single cases that can be memorized, our questions are sequence-based, allowing numerous cases for validating true understanding. 
(2) General Solutions. UTMath requires large models to solve problems by generating code, aiming for general solutions rather than problem-specific ones, reflecting a closer alignment with intelligence.

\vspace{-5pt}
\subsection{Building Methods}
Constructing effective, high-quality datasets is a complex and labor-intensive process. The advent of LLMs offers an opportunity to change this scenario ~\cite{valmeekam2022large, drori2023human, perez2022discovering, chiang2023can, liu2023g, fu2023chain, kocmi2023large,li2024crowdsourced}. For instance, ~\cite{almoubayyed2023rewriting} employed GPT-4o to rewrite mathematics problems based on MATHia ~\cite{ritter2007cognitive} to aid students in improving their math performance, validating this with 12,374 students and demonstrating the effectiveness of using LLMs for data construction. These efforts provide a reliable foundation for utilizing LLMs in data processing. 

In our study, we utilized GPT-4o to help us deal with data, such as by providing necessary background knowledge for questions and making them more understandable, with more information about the prompts used shown in the Appendix \ref{sec:appendix:prompts}. Subsequently, human verification was performed to ensure consistency before and after LLM usage.

\subsection{Prompting Methods}
Considering the attributes of large models, they exhibit significant sensitivity to prompts, rendering prompt engineering a critical area of study. 

The Chain-of-Thought~\cite{wei2022chain} prompting technique encourages models to express reasoning steps in natural language before concluding. Similarly, the approach by ~\cite{kojima2022large} uses the phrase "Let's think step by step" to effectively guide large language models through their reasoning.
Inspired by CoT, several effective prompting methods have been developed, such as Tree-of-Thoughts~\cite{yao2024tree}, Graph-of-Thoughts~\cite{besta2024graph}. .
Program-of-Thought prompting~\cite{chen2022program}: PoT generates programs as the intermediate steps and integrates external tools like a Python interpreter for precise calculations, as well as other prompting methods~\cite{wang2023plan, gao2023pal}.

Our RCoT method stands out by dividing reasoning into two steps: reasoning and implementing based on reasoning. This segmentation provides deeper insights into LLM reasoning abilities. The advantages can be summarized as follows. (1) Enhanced Reasoning. Emphasizing reasoning allows large models to focus more on improving the quality of reasoning, thereby delivering higher-quality and more efficient solutions.
(2) Modularity. By separating reasoning from implementation, the influence of coding on reasoning can be eliminated, providing a new paradigm for evaluating the reasoning ability through the code generated by the model.

\begin{figure*}
  \centering
  \includegraphics[width=1\textwidth]{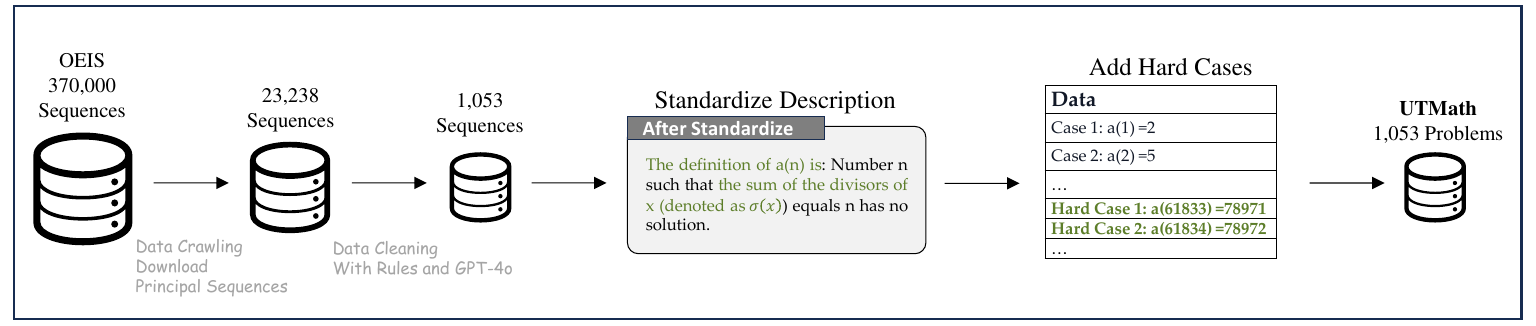}
  \caption{UTMath generation pipeline.
  After downloading 23,238 Principle Sequences from OEIS and cleaning the data, 1,053 usable sequences were obtained. Descriptions were standardized by adding background information and improving readability (highlighted in green, also shown in Appendix~\ref{sec:appendix:B2}). Hard cases were introduced to enhance discriminative capability, including terms from later positions to prevent simplistic algorithms from passing.}
  \label{fig:UTMath generation pipline}
\end{figure*}

\section{UTMath Benchmark}
\subsection{Introduction for OEIS.}
The OEIS was established to document integer sequences of interest to both professional and amateur mathematicians, and it has become widely cited in the community. 
Most sequences are derived or updated from academic papers, contributing to their cutting-edge level of difficulty.
As of February 2024, it contains over 370,000 sequences~\cite{OEIS2024}. 
Each sequence is accompanied by an identification number, a brief description, some sample integers, links to relevant literature, and, where possible, program code for computing the sequences. An example sequence is shown in Appendix~\ref{sec:appendix:oeis_sequence}.

\subsection{Benchmark Construction.}
\label{sec:benchmark_construction}
UTMath is a cutting-edge and expansive benchmark designed to more accurately assess the mathematical reasoning abilities of LLMs. It consists of 1053 math problems, with each problem having an average of 68 test cases. The benchmark covers 9 mathematical domains, including not only common topics like number theory but also graph theory, group theory, topology, and geometry. The difficulty of UTMath is considered Cutting-Edge, with the majority of the sequences that form the problems having been studied in academic papers. UTMath was obtained as follow (see also Fig.\ref{fig:UTMath generation pipline}).

\paragraph{Data Crawling.}
OEIS provides users with a list of principal sequences~\footnote{\url{https://oeis.org/wiki/Index_to_OEIS}}, which are most important sequences defined by OEIS. OEIS categorizes these sequences into sections based on the first 2-3 letters of their content themes. 
By scraping the category tags within each section and the AIDs of their subordinate sequences, we obtained 23,238 principal sequences' AIDs. 
OEIS provides an interface to request the JSON data of the HTML page for each sequence using its AID~\footnote{\url{https://oeis.org/wiki/JSON_Format}}. By passing the sequence AIDs to this interface, we acquired the JSON data for these 23,238 sequences.

\paragraph{Data Cleaning.}
We found that some of the sequences we collected did not meet our criteria and should be removed, with further details provided in the Appendix~\ref{sec:appendix:data_cleaning}. Here are several main situations.

$\bullet$ \textbf{Hard to solve, few terms are discoverable.} A portion of the sequences retrieved are marked as ``hard'' in the keyword field of their entries in OEIS. According to OEIS, ``Any sequence which can be extended only by new ideas, rather than more computation deserves keyword: hard. Similarly, if computing a term of the sequence would probably merit a paper in a peer-reviewed journal (discussing the result, the algorithm, etc.)''
\footnote{\url{https://oeis.org/wiki/User:Charles_R_Greathouse_IV/Keywords/difficulty}}
Another related keyword attribute is "fin" (finite), indicating sequences with limited length. For our purposes, sequences should be infinitely derivable.
 
$\bullet$ \textbf{Difficult to Generate Programmatically.} In OEIS, most sequences are provided with fields such as Mathematica, program, or formula, but not all sequences include these details. We assume that the sequences without these fields may be difficult to generate programmatically.

$\bullet$ \textbf{Simple Sequences.} 
Some sequences are too simple to require any reasoning. We use GPT-4o to determine if a sequence requires reasoning or just implementation; if mostly implementation, it's excluded. 
For instance, A000178\footnote{\url{https://oeis.org/A000178}}: 'Superfactorials: product of the first n factorials,' a sequence requiring only implementation, will be excluded.

After addressing the aforementioned issues, we ultimately obtained 1053 sequences.

\definecolor{darkred}{rgb}{0.5, 0, 0}
\begin{table*}
  \centering
  \small
  \begin{tabular}{lcccccc}
  \toprule
  \textbf{Dataset} & \textbf{Size} & \textbf{Level} &  \textbf{Multi-test} & \textbf{Efficiency} & \textbf{Metric} & \textbf{Output} \\ \midrule
  College Math  & 2,818 & University &  \textcolor{darkred}{\ding{55}} & \textcolor{darkred}{\ding{55}} & Accuracy & Text\\
  GSM8K & 1,319 & Elementary school &  \textcolor{darkred}{\ding{55}} & \textcolor{darkred}{\ding{55}} & Accuracy & Text\\
  MATH & 5,000 & High school & \textcolor{darkred}{\ding{55}} & \textcolor{darkred}{\ding{55}} & Accuracy & Text\\ 
  RobustMath & 300 & High school & \textcolor{darkred}{\ding{55}} & \textcolor{darkred}{\ding{55}} & Accuracy & Text\\ 
  OlympiadBench & 8,476 & Competition &  \textcolor{darkred}{\ding{55}} & \textcolor{darkred}{\ding{55}} & Accuracy & Text\\
  TheoremQA & 800 & University &  \textcolor{darkred}{\ding{55}} & \textcolor{darkred}{\ding{55}} & Accuracy & Text\\
  \midrule
  UTMath(ours) & 1,053 & Cutting-edge &  \textcolor{darkgreen}{\ding{51}} & \textcolor{darkgreen}{\ding{51}} & Pass Rate & Code\\
  \midrule
  \end{tabular}
\caption{Comparison between UTMath and other benchmarks. UTMath offers a cutting-edge benchmark with a comprehensive set of 1,053 problems across multiple mathematical domains, providing a more accurate evaluation of LLMs' mathematical reasoning capabilities.}
  \label{tab:Comparison between UTMath and other benchmarks}
\end{table*}

\paragraph{Standardization of Question Statements.}

As a specialized academic database in the field of mathematics, OEIS provides a wealth of useful information for each sequence. However, we have found that some sequences cannot be directly used with the descriptions provided by OEIS as problem statements, primarily for the following reasons: (1) Specialized Terminology. Some sequence descriptions use complex math terms that need examples or explanations to be clear. Using them directly as problems might test mathematical knowledge rather than reasoning skills. So, it is important to explain key concepts to focus on reasoning and reduce the extra knowledge needed.(2) Brevity and Ambiguity. Some sequence descriptions are excessively brief and lack a clear definition of what a(n) is. 
We used GPT-4o to standardize these by adding background info and making the language smoother. The prompts we used are provided in the Appendix \ref{sec:appendix:prompts} and an example is shown in Appendix \ref{sec:appendix:B2}. Despite our requirement that ensuring the consistency of meaning between the original and processed descriptions, hallucinations can still occur. To mitigate this, we performed a manual verification of the standardized problem statements to ensure they matched the original descriptions in meaning and were easy to comprehend.

\paragraph{Hard Test Cases Mining.}

The primary goal of this paper is to evaluate the reasoning ability of LLMs. Generally, more efficient solutions to a problem imply stronger reasoning capabilities. Therefore, we aim for our evaluation to distinguish whether a solution is efficient.
However, in the OEIS, each sequence only lists the first few n terms, normally n<100, which can be obtained without requiring particularly efficient methods. This limitation prevents the evaluation from effectively distinguishing between efficient and inefficient solutions. An obvious fact is that the difficulty of computing the first 10 terms of a sequence within a time limit is significantly different from computing terms starting from the $10^6$th term. Therefore, we aim to create more challenging test data to better assess the reasoning capabilities of LLMs.

Fortunately, many OEIS sequences include corresponding Mathematica code that can be regarded as the ground-truth solution for each problem. We extract this Mathematica code for each sequence, formalizing it to compute the first $N$ terms, $A_1, ..., A_N$, of the sequence. We determine the maximum value of $N_{max}$ for which the code can compute the sequence within 10 seconds, where we set $10^6$ as the upper bound. Finally, we add the last 10 terms $A_{N_{max-9}}, ..., A_{N_{max}}$ into our benchmark as the hard test cases to evaluate the complexity of a solution. 
Our experiments demonstrate that these cases precisely differentiate more efficient and intelligent solutions.

\subsection{Evaluation Metrics}
We adopt the metric pass@$k$ to evaluate the performance of LLMs. 
The metric pass@$k$ is a classic metric in code generation, where a problem is solved if any of the 
k generated samples passes the unit tests. We use the stable method of calculation proposed by~\citep{chen2021evaluating}:

\vspace{-10pt}

\begin{equation}
    \text{pass@}k := \mathbb{E}_{\text{Problems}} \left[ 1 - \binom{n-c}{k} \big/ {\binom{n}{k}} \right]
\end{equation}

\begin{table}
  \centering
  \small
  \begin{tabular}{lc}
  \toprule
  \textbf{Category} & \textbf{\# of Problems}\\ \midrule
  Number Theory & 159\\ 
  Graph Theory  & 79\\
  Group Theory  & 65\\
  Discrete Mathematics & 158\\
  Combinatorial Mathematics & 158\\
  Geometry and Topology & 70\\
  Poly. and Series Expan. & 151\\ 
  Special Numbers & 157\\
  Formal Languages & 56\\
  \midrule
  \textbf{Total} & 1053\\ 
  \bottomrule
  \end{tabular}

\caption{Categories and distribution of problems.}
  \label{tab:Categories and distribution of problems}
\end{table}

\definecolor{lightergray}{rgb}{0.9, 0.9, 0.9}
\definecolor{darkred}{rgb}{0.6, 0.1, 0.1}
\begin{table*}[h]
    \centering
    \small
    \begin{tabular}{l|c>{\columncolor{lightergray}}c|c>{\columncolor{lightergray}}c|cc>{\columncolor{lightergray}}r}
         \toprule
         \textbf{Model} & \multicolumn{2}{c|}{\textbf{Pass@1 (\%) $\uparrow$}} & \multicolumn{2}{c|}{\textbf{Pass@5(\%) $\uparrow$}} & \multicolumn{3}{c}{\textbf{Avg. Run Time (s) $\downarrow$}}\\
         & \textbf{PoT} & \textbf{RCoT}& \textbf{PoT} & \textbf{RCoT} & \textbf{PoT} & \textbf{RCoT} & \textbf{Efficiency}  \\
         \midrule
         \multicolumn{8}{c}{\textit{\textcolor{gray}{closed-source models}}} \\
         o1-mini & \textbf{29.34} & \textbf{32.57} {\footnotesize \textcolor{darkgreen}{($+$3.23)}} & ------ & ------ & 5.58 & 3.76 & \textcolor{darkgreen}{$+$32.62\%} \\
         o1-preview & 23.74 & 27.16 {\footnotesize \textcolor{darkgreen}{($+$3.42)}} & ------ & ------ & 4.66 & 3.96 & \textcolor{darkgreen}{$+$15.02\%} \\
         GPT-4o  & 25.53 & 26.93 {\footnotesize \textcolor{darkgreen}{($+$1.40)}} & \textbf{32.67} & \textbf{35.90} {\footnotesize \textcolor{darkgreen}{($+$3.23)}} & 6.98 & 6.23 & \textcolor{darkgreen}{$+$12.04\%} \\
          Gemini-1.5-Pro & 19.70 & 19.43 {\footnotesize \textcolor{darkred}{($-$0.27)}} & 31.24 & 33.14 {\footnotesize \textcolor{darkgreen}{($+$1.90)}}  & 6.30 & 6.22 & \textcolor{darkgreen}{$+$1.28\%} \\
         Claude-3.5-Sonnet & 18.58 & 19.11 {\footnotesize \textcolor{darkgreen}{($+$0.53)}} & 27.83 & 31.34 {\footnotesize \textcolor{darkgreen}{($+$3.51)}} & 6.44 & 5.32 & \textcolor{darkgreen}{$+$21.05\%} \\
         GPT-3.5-Turbo & 11.68 & 6.82 {\footnotesize \textcolor{darkred}{($-$4.86)}} & 17.09 & 13.30 {\footnotesize \textcolor{darkred}{($-$3.79)}}  & 5.42 & 5.06 & \textcolor{darkgreen}{$+$7.11\%} \\
         \midrule
         \multicolumn{8}{c}{\textit{\textcolor{gray}{open-source models}}} \\
         Qwen2.5-72B & \textbf{23.48} & \textbf{22.17} {\footnotesize \textcolor{darkred}{($-$1.31)}} &  \textbf{31.05} & \textbf{33.33} {\footnotesize \textcolor{darkgreen}{($+$2.28)}} & 5.88 & 4.31 & \textcolor{darkgreen}{$+$36.42\%} \\
         DeepSeek-V2.5-236B & 20.95 & 21.63 {\footnotesize \textcolor{darkgreen}{($+$0.68)}} & 30.10 & 31.72 {\footnotesize \textcolor{darkgreen}{($+$1.62)}}  & 6.64 & 5.44 & \textcolor{darkgreen}{$+$22.06\%} \\
         Qwen2.5-Math-72B & 19.72 & 20.53 {\footnotesize \textcolor{darkgreen}{($+$0.81)}} & 26.69 & 28.11 {\footnotesize \textcolor{darkgreen}{($+$1.42)}}  & 5.04 & 3.81 & \textcolor{darkgreen}{$+$24.40\%} \\
         LLaMA-3.1-405B & 15.76 & 16.09 {\footnotesize \textcolor{darkgreen}{($+$0.33)}} & 25.26 & 27.35 {\footnotesize \textcolor{darkgreen}{($+$2.09)}}  & 5.73 & 5.12 & \textcolor{darkgreen}{$+$11.91\%} \\
         \bottomrule
    \end{tabular}
    \caption{Pass Rate and Average Run Time of LLMs on UTMath. We listed the performance of eight large models by the PoT or the RCoT methods across a range of metrics. For o1-mini and o1-preview only Pass@1 data is currently available due to resource constraints. The average run time is calculated based on the problems solved by both the PoT and the RCoT methods. 
    The efficiency is calculated as: (Avg.Runtime(PoT) - Avg.Runtime(RCoT)) / Avg.Runtime(RCoT). Two qualitative cases are shown in Appendix \ref{sec:appendix:Case Study}. 
    }
    \label{tab:Evaluation on UTMath}
\end{table*}

\subsection{Dataset Statistics}
 The main statistics of UTMath are shown in Tab.~\ref{tab:Comparison between UTMath and other benchmarks}. To gain a deeper understanding of the composition of the UTMath Benchmark, we identified nine mathematical fields and used GPT-4o to categorize each problem to these fields as shown in Tab.~\ref{tab:Categories and distribution of problems}. 
Our analysis reveals that only 10 out of 1,053 problems have no references. The reference years span from 1950 to 2024, with the maximum number of references exceeding 6,000. These findings underscore the cutting-edge nature of our benchmark. More details can be found in Appendix~\ref{sec:appendix:data_cleaning}.

\section{Reasoning-to-Coding of Thoughts}
Compared to methods that simply check whether the outputs generated by LLMs are identical, the code-based evaluation approach enables more accurate assessment by using multiple test cases. It provides additional evaluation metrics, such as runtime, and allows for the observation of the reliability of LLMs.

Initially, we adopted the Program of Thought (PoT) method, where the LLM had to perform reasoning and implement it in one step. However, we noticed that the LLMs often resorted to simpler algorithms, which led to high time complexity or even failure with more complex problems due to limited reasoning depth. To improve this, we explored the \textbf{Reasoning-to-Coding of Thoughts (RCoT)} framework, which separates reasoning and implementation into different steps.

In the first round, LLM focuses only on reasoning about the problem. Compared to PoT, RCoT dedicate an entire round to reasoning allowing the LLM to generate a step-by-step, detailed logical reasoning chain, including mathematical theorems, formulas, and properties used. This deeper reasoning approach facilitates the creation of more efficient algorithms with lower time complexity.

In the second round, RCoT requires the LLM to implement the reasoning process generated in the first round. By converting the computation process into code, we can introduce new metrics beyond the final pass rate. The solution generated by the LLM can be evaluated based on its runtime during testing, which indirectly reflects the solution's time complexity. Additionally, this approach avoids errors caused by the limited computational capabilities of large models, allowing for more accurate and genuine insights into the LLM’s reasoning abilities. Clearly, the stronger the reasoning capabilities of the large model, the higher the overall pass rate and the lower the time complexity, which is reflected in the shorter runtime of the generated solution.

We present qualitative cases in Appendix~\ref{sec:appendix:Case Study}, where GPT-4o solves problems from UTMath using PoT and RCoT, respectively.
The case studies show that, with RCoT prompting, the model engages in deeper reasoning, significantly reducing solution complexity.

\section{Experiment}
\label{sec:exp}

\begin{figure*}
  \centering
  \includegraphics[width=0.9\textwidth]{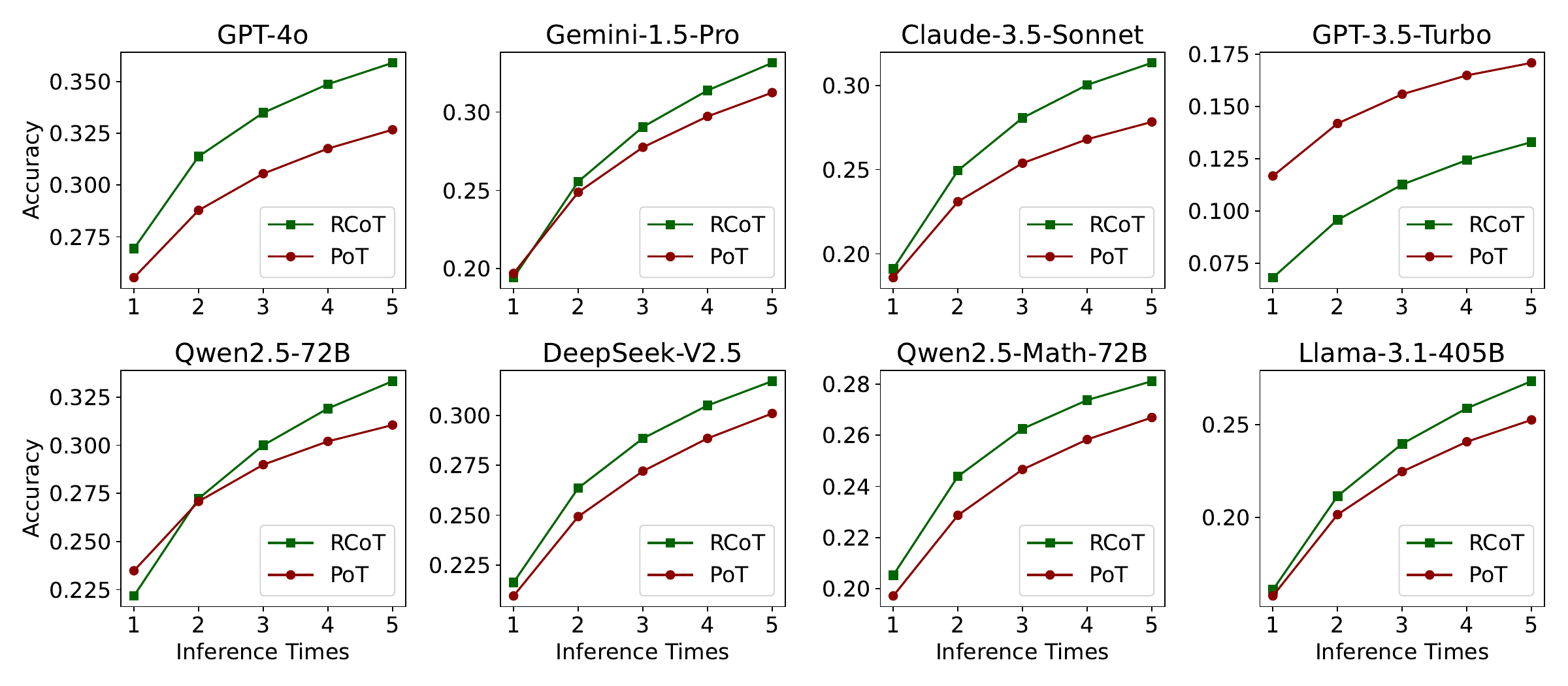}
  \caption{Performance comparison of models across PoT and RCoT tasks at different pass@k levels.}
  \label{fig:exp_passk}
\end{figure*}

\subsection{Experimental Setup}

Here, we consider the closed-source models, i.e., GPT-3.5-Turbo, GPT-4o, o1-mini and o1-preview from OpenAI~\cite{GPT4o}, Claude-3.5-Sonnet~\cite{Claude3.5}, Gemini-1.5-Pro~\cite{reid2024gemini}, as well as the open-source models, i.e., LLaMA-3.1~\cite{dubey2024llama}, Qwen2.5~\cite{Qwen2.5}, Qwen2.5-Math~\cite{Qwen2.5-Math}, DeepSeek-V2.5~\cite{bi2024deepseek}. The metric pass$@1$ is calculated as the average result over 5 run times. 
We run all evaluations in a laptop with CPU Intel(R) Core(TM) i7-10750H CPU @ 2.60GHz.

\subsection{Evaluation on UTMath}

Here we evaluate both open-source and closed-source models using \OurP and PoT in Tab.~\ref{tab:Evaluation on UTMath}. 
The experimental results shows that all tested models performed poorly on our benchmark.
The best model, o1-mini, only solves 32.57\% problem in our benchmark, followed by o1-preview at 27.16\% and GPT-4o at 26.93\%. 
Since our problems are sourced from the OEIS, they consist of sequences and solutions proposed by various mathematicians in the context of cutting-edge research.
This suggests that our benchmark is challenging enough to help guide future directions for improving LLMs.

Compared to PoT, our method \OurP demonstrates superiority in two aspects. First, prompting with \OurP achieves higher pass$@5$ performance across 7 LLMs, with the best results observed on GPT-4o. Second, the solutions generated by \OurP for all LLMs demonstrate more efficient performance, particularly Qwen2.5-72B, where the RCoT approach achieves an efficiency improvement of over 36.42\% compared to PoT, as shown in Tab.~\ref{tab:Evaluation on UTMath} and Fig.~\ref{fig:RCoT_results}. It indicates that, RCoT prompting enables the model to engage in deeper reasoning, significantly reducing solution complexity and enhancing solution performance.
However, some models experienced a decrease in pass$@1$ with \OurP. 
Specifically, the accuracies of Gemini-1.5-Pro, GPT-3.5-Turbo, and Qwen2.5-72B slightly dropped. Notably, while Gemini-1.5-Pro and Qwen2.5-72B experienced a drop in pass$@1$, their pass$@5$ performance improved. It indicates that \OurP brings more room in multiple inference times.
The observed decrease in performance may stem from the fact that formulating more efficient solutions often requires higher-level reasoning, which can increase the difficulty of the task and make these models more susceptible to errors when attempting more sophisticated solutions.

\begin{table}
  \centering
  \small
  \begin{tabular}{clrr}
  \toprule
  & \textbf{Model} & \textbf{Easy} & \textbf{Easy \& Hard} \\
  \midrule
  \multirow{4}{*}[0px]{\rotatebox{90}{closed}} & GPT-4o & \textbf{34.95} & \textbf{26.93} \\
  & Gemini-1.5-Pro & 23.84 & 19.43 \\ 
  & Claude-3.5-Sonnet & 24.86 & 19.11 \\
  & GPT-3.5-Turbo & 8.72 & 6.82 \\
  \midrule
  \multirow{4}{*}[0px]{\rotatebox{90}{open}} & Qwen2.5-72B & \textbf{28.96} & \textbf{22.17} \\ 
  & DeepSeek-V2.5 & 27.52 & 21.63 \\ 
  & Qwen2.5-Math-72B & 24.60 & 20.53 \\
  & LLaMA-3.1-405B & 22.09 & 16.09 \\
  \bottomrule
  \end{tabular}
    \caption{Performance (\%) of different models on easy and hard test cases. 
    Easy cases: The initial terms in OEIS. Hard cases: mined hard test cases (\S~\ref{sec:benchmark_construction}).
    }
  \label{tab:performance on hard case}
\end{table} 

\begin{table*}
  \centering
  \small

  \begin{tabular}{l|ccccccccc|c}
  \toprule
  \textbf{Model} & NT & Graph T. & Group T. & DM & CM & GT & PSE & SN & FL  & \textbf{pass@1} \\
  \hline
  \multicolumn{11}{c}{\textit{\textcolor{gray}{closed-source models}}} \\
o1-mini  & \textbf{52.83} & \textbf{7.59} & \textbf{15.38} & \textbf{42.41} & \textbf{32.27} & \textbf{7.14} & \textbf{23.84} & \textbf{40.13} & 37.50 & \textbf{32.57}\\

o1-preview  & 47.17 & 6.33 & 13.85 & 34.17 & 25.32 & 2.86 & 23.18 & 29.94 & 33.93 & 27.16\\

GPT-4o  & 43.90 & 2.78 & 11.69 & 38.23 & 24.94 & 3.43 & 16.42 & 33.89 & \textbf{42.50} & 26.93\\

Gemini-1.5-Pro  & 31.70 & 1.27 & 8.92 & 27.47 & 15.19 & 5.71 & 15.23 & 27.39 & 17.86 & 19.43\\

Claude-3.5-Sonnet & 33.58 & 1.52 & 8.00 & 29.49 & 12.91 & 5.43 & 11.52 & 26.62 & 20.36 & 19.11\\

GPT-3.5-Turbo & 13.08 & 0.00 & 1.85 & 11.39 & 3.29 & 0.29 & 2.78 & 10.96 & 8.93 & 6.82\\
\hline
\multicolumn{11}{c}{\textit{\textcolor{gray}{open source models}}} \\
Qwen2.5-72B & 36.86 & \textbf{2.53} & \textbf{12.00} & 30.63 & 15.95 & \textbf{6.00} & \textbf{18.15} & 29.43 & \textbf{24.29} & \textbf{22.17}\\
DeepSeek-V2.5  & \textbf{38.24} & 1.27 & 8.92 & \textbf{33.16} & \textbf{17.34} & 2.29 & 12.45 & \textbf{31.08} & 20.00 & 21.63\\
Qwen2.5-Math-72B  & 35.35 & 1.27 & 8.62 & 28.73 & 14.81 & 4.00 & 17.48 & 28.15 & 20.00 & 20.53 \\
LLaMA-3.1-405B & 29.56 & 0.76 & 4.92 & 25.44 & 9.62 & 2.00 & 9.54 & 22.55 & 21.43 & 16.09\\

\bottomrule
  \end{tabular}

  \caption{
  Performance (\%) on different problem categories. 
  Categories are represented by abbreviations. NT: Number Theory; T.: Theory; DM: Discrete Mathematics; CM: Combinatorial Mathematics; GT: Geometry and Topology; PSE: Polynomial and Series Expansions; SN: Special Numbers; FL: Formal Languages.
  }
  \label{tab:performance on different categories}
\end{table*}

\subsection{The Effectiveness of Hard Test Cases}
As we mentioned in \S~\ref{sec:benchmark_construction}, each sequence in the OEIS lists only the initial terms, which we refer to as ``easy test cases''. 
To investigate the model's ability to handle challenging cases, we evaluated whether it could predict values that appear later (i.e., $10^6$) in a sequence. 
These later values are typically underrepresented in pre-training data and often require more computation time and a more precise implementation to retrieve accurately.
The experimental results, which depicted in Tab.~\ref{tab:performance on hard case}, reveal that the model's performance drops significantly when handling these hard cases. 
This indicates that introducing these cases can prevent simple solutions from passing all the test cases, thereby filtering for more advanced solutions.

\subsection{Scaling of the Inference Times}
\label{sec:exp_pass@k}
We compared the performance difference between running the LLMs five times and reported the metric of pass$@k$. 
As shown in Fig.~\ref{fig:exp_passk}, all models improved their performance with an increasing number of inference times. 
For Qwen2.5-72B and Gemini-1.5-Pro, RCoT was slightly weaker than PoT in pass$@1$ but quickly approached and surpassed PoT in subsequent run times. 
We observed that with an increasing number of inference time, RCoT consistently demonstrated a growing advantage in performance across almost all models, except for GPT-3.5. However, it is worth noting that GPT-3.5 exhibited the lowest pass rate. 
This suggests that RCoT may perform better in models with stronger reasoning capabilities.

\subsection{Importance of the Reasoning Step}
\label{sec:exp_reasoning}
\begin{figure}
  \centering
  \includegraphics[width=0.5\textwidth]{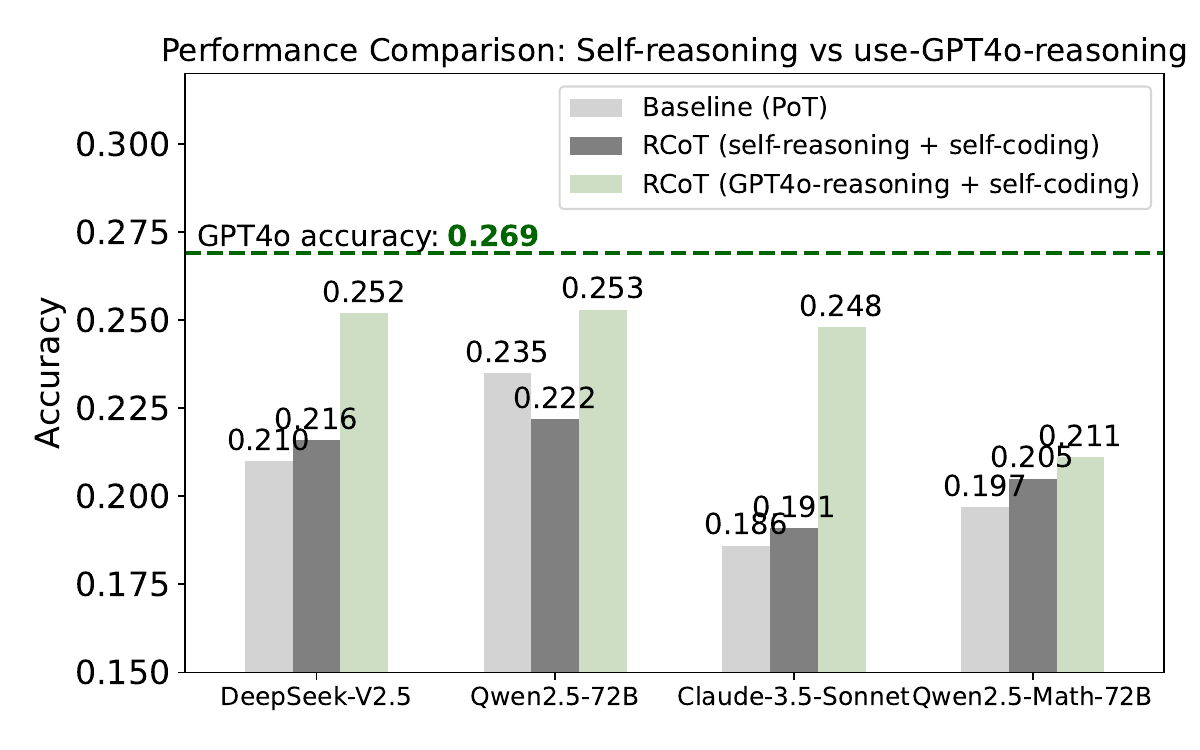}
  \caption{Performance comparison between self-reasoning and using GPT-4o reasoning for coding across different models. The results show that models perform better when relying on GPT-4o's reasoning output.}
  \label{fig:exp_reasoning_comparison}
\end{figure}

GPT-4o has the best performance, while we are unclear whether this was due to its superior reasoning or coding ability.
To investigate further, we tested using GPT-4o for the reasoning step while other models perform the coding step based on the reasoning result from GPT-4o.
As depicted in Fig.\ref{fig:exp_reasoning_comparison}, the results showed that the performance of models increased significantly when implementing coding based on GPT-4o's reasoning output, 
suggesting that the reasoning quality is important and GPT-4o does produce higher-quality reasoning results.

\subsection{Performance on Different Categories}
Our benchmark comprehensively evaluates the LLMs' ability across various categories of math problems. 
GPT-4o achieved the highest score in the formal language domain, while o1-mini achieved the best scores in the remaining eight domains.
All models performed very poorly in the categories of Graph Theory, Group Theory, and Geometry and Topology, with accuracy rates below 16\%, highlighting the need for further exploration in these areas. 

\section{Conclusion}
In this work, we investigate how to more accurately and effectively evaluate the mathematical reasoning capabilities of LLMs. We propose a cutting-edge benchmark, UTMath, which comprises 1,053 problems spanning nine mathematical domains, with an average of 68 test cases per problem. 
This benchmark presents significant challenges: o1-mini, the best-performing model, successfully solves only 32.57\% of the problems, followed by o1-
preview at 27.16\%, and GPT-4o at 26.93\%. 
Additionally, we introduce RCoT (Reasoning-to-Coding of Thought). Our study finds that, compared to PoT (Program-of-Thought), RCoT significantly improves the algorithmic efficiency and pass rates of most models by facilitating deeper reasoning steps prior to code generation. 
Overall, this research contributes to a deeper understanding of the current capabilities of LLMs in mathematical reasoning and lays the groundwork for the development of more advanced models in the future.
\section*{Limitation}
The primary limitation of UTMath lies in the evaluation metrics: the performance of the evaluation machine affects the runtime of the generated code, making the absolute numerical results incomparable across different machines. We utilized an i7-10750H processor to execute the reference solutions and conduct evaluations, and we recommend using the same machine for testing and replication.
There are two main limitations of RCoT. First, we only installed a set of common packages, such as sympy, in the standard testing environment. This avoids allowing LLMs to call highly integrated packages while also preventing the generation of potentially harmful code that could damage the evaluation system. Second, while our experiments demonstrate the critical role of reasoning quality in determining success rates, we have not further explored methods for enhancing reasoning quality, which remains an area for future investigation.

\section*{Ethics Statements}

The UTMath Benchmark is designed to advance the evaluation of mathematical reasoning in LLMs. We recognize the potential ethical concerns associated with this work, particularly the risk of data misuse. To mitigate this, we strictly adhere to usage guidelines and licensing terms for the UTMath-Train dataset, which is intended solely for academic and research purposes.
While the UTMath Benchmark evaluates model performance in terms of accuracy and generality, automated evaluations may introduce biases due to the nature of the datasets and evaluation algorithms. Additionally, while UTMath covers a wide range of mathematical domains, it may not fully represent diverse cultural or educational perspectives. We encourage further development of benchmarks that incorporate a broader array of reasoning styles to ensure more inclusive evaluations.
By releasing UTMath, we aim to foster responsible AI development, promoting better, more generalizable mathematical reasoning systems.

\bibliography{custom}

\newpage
\onecolumn 
\appendix

\section{An Example Sequence in OEIS}

\label{sec:appendix:oeis_sequence}
\begin{figure}[H]
    \centering
    \includegraphics[width=1\linewidth]{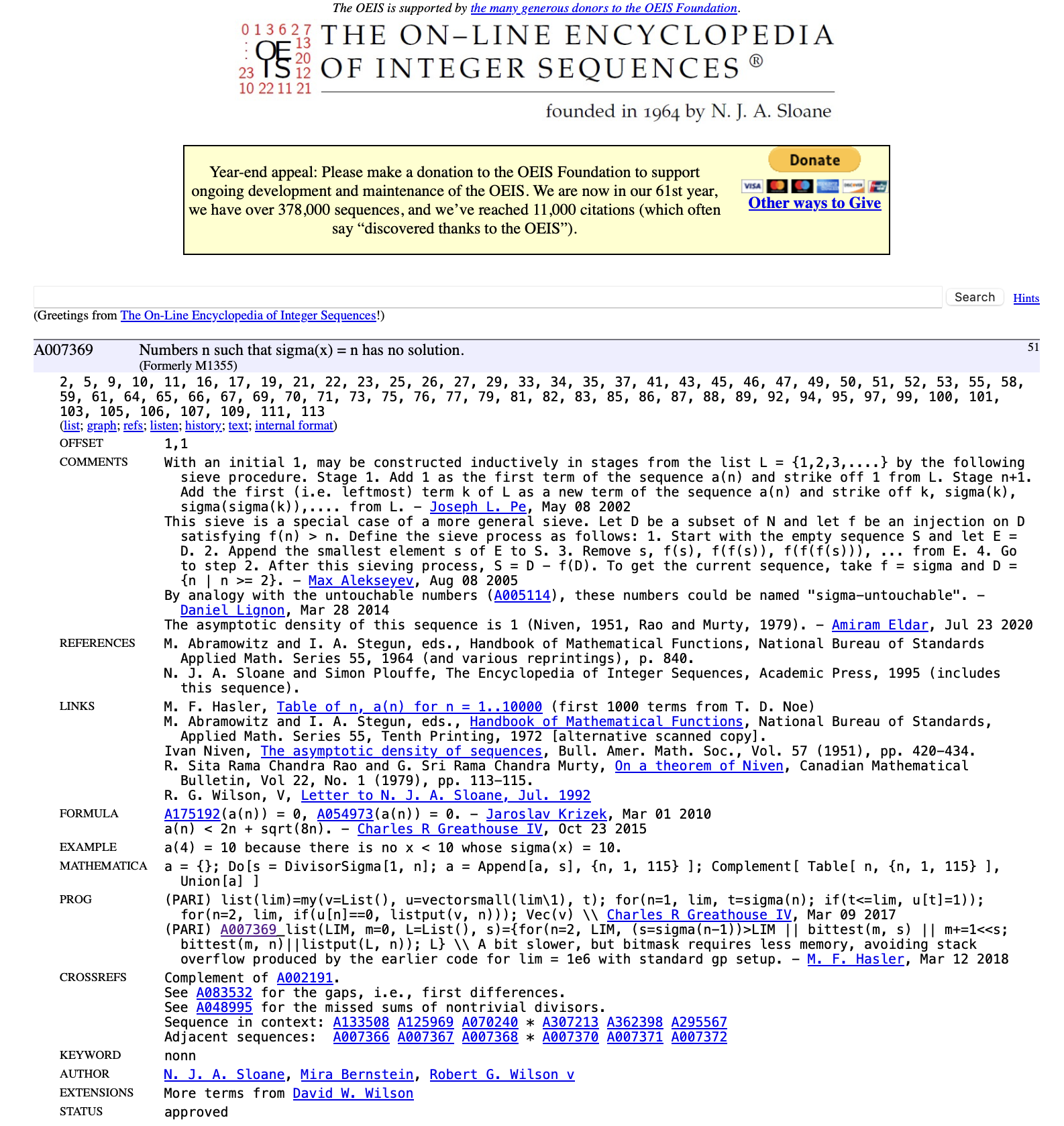}
    \caption{Sequence A007369 in OEIS. Its description: "Numbers n such that sigma(x) = n has no solution." (Clearly, without specific background knowledge, we cannot fully understand what the function sigma() represents, which is one of the reasons we perform standardization. \S\ref{sec:appendix:B2}) Next, OEIS shows the first 67 terms of this sequence, which we classify as easy cases. Below that, additional metadata is provided, including comments, references, links, formulas, examples, programs, author, status, and more. It is evident that this sequence has garnered significant attention from researchers, reflecting the Cutting-Edge difficulty of our benchmark. We used the Mathematica program included in the metadata to generate Hard cases, with detailed procedures provided in \S~\ref{sec:benchmark_construction}. As a scientific database, each sequence submitted to OEIS undergoes a review process, and the status "approve" indicates that the sequence has been validated and approved by OEIS administrators.}
    \label{fig:oeis_sequence}
\end{figure}
\twocolumn
\clearpage
\section{Dataset Construction Details}
\label{sec:appendix:data_cleaning}
This section primarily presents some details on the construction of UTMath. ~\ref{sec:appendix:B1} discusses the issues encountered when observing data crawled from OEIS, along with the corresponding cleaning rules. UTMath applies all 14 rules. Additionally, we crawled all sequences from OEIS and, for convenience, applied only the first 12 rules to create UTMath\_Train, which contains over 70k sequences. ~\ref{sec:appendix:B2} outlines the process followed for standardizing the descriptions of problems in UTMath, while
~\ref{sec:appendix:B3} explains the referencing of sequences within UTMath, highlighting both the Cutting-Edge difficulty level of UTMath and its scalability.

\subsection{Rules for Data Cleaning}
\label{sec:appendix:B1}
\hspace{1em}
\begin{spacing}{0.3}  
\vspace{-20pt}
\begin{enumerate}[itemindent=0em, itemsep=-0.2em]
    \begin{spacing}{0.75}
    \item Issues: The sequence is too difficult, requiring extensive background knowledge, or only a limited number of terms are found. 
    \vspace{-4pt}
    
    Method: Remove sequences with keywords containing ‘hard’, ‘fin’ (finite).
    \end{spacing}
    \vspace{-5pt}
    \begin{spacing}{0.75}
    \item Issues: The sequence is hard to generate with a program.
    \vspace{-4pt}
    
Method: Check if it contains program, formula, or Mathematica fields in the sequence's json data.
\end{spacing}
\vspace{-5pt}
    \begin{spacing}{0.75}
    \item Issues: The sequence is too simple with an explicit recurrence or closed formula.
    \vspace{-5pt}
    
Method: Search if the description includes ‘a(n) =’.
\end{spacing}
\vspace{-5pt}
    \begin{spacing}{0.75}
    \item Issues: Solving the sequence requires information from other OEIS sequences.
    \vspace{-4pt}
    
Method: Search if the sequence's description contains the AID of other sequences (‘A’ + six-digit number with leading zeros).
\end{spacing}
\vspace{-5pt}
    \begin{spacing}{0.75}
    \item Issues: The sequence is decimal expansion of a certain number.
    \vspace{-5pt}
    
Method: Search if the description includes ‘decimal’.
\end{spacing}
\vspace{-5pt}
    \begin{spacing}{0.75}
    \item Issues: The sequence consists of repetitions or a constant value.
    \vspace{-5pt}
    
Method: Search if the description includes both 'repeat' and 'period' or 'constant sequence'.
\end{spacing}
\vspace{-5pt}
    \begin{spacing}{0.75}
    \item Issues: The description is too vague.
    \vspace{-4pt}
    
Method: Search if the description includes ‘related to’.
\end{spacing}
\vspace{-5pt}
    \begin{spacing}{0.75}
    \item Issues: Another version of a concept.
    \vspace{-5pt}
    
Method: Search if the title includes ‘another version’, ‘second kind’, etc.
\end{spacing}
\vspace{-5pt}
    \begin{spacing}{0.75}
    \item Issues: The sequence is formed by taking mod of a constant.
    \vspace{-4pt}
    
Method: Search if the description includes ‘module’.
\end{spacing}
\vspace{-5pt}
    \begin{spacing}{0.75}
    \item Issues: The values in the sequence are too large, which might cause LLM tokenization errors. 
    \vspace{-5pt}
    
Method: Check if any term’s length exceeds 18 digits(i.e., greater than $1e18$), remove it.
\end{spacing}
\vspace{-5pt}
    \begin{spacing}{0.75}
    \item Issues: Coefficient triangles or ‘read by row’ topics.
    \vspace{-5pt}
    
Method: Search if the sequence's description includes ‘read by row’, ‘triangle of coefficient’.
    \end{spacing}
    \vspace{-5pt}
    \begin{spacing}{0.75}
    \item Issues: The description is too short, either purely implementation or lacks necessary information.
    \vspace{-5pt}
    
Method: Check if the title length is below 5.
    \end{spacing}
    \vspace{-5pt}
    \begin{spacing}{0.9}
    \item Issues: More like a reasoning puzzle.
    \vspace{-5pt}
    
Method: Use GPT-4o to judge, with the prompt outlined in the Appendix~\ref{sec:appendix:prompts}.
    \end{spacing}
    \vspace{-5pt}
    \begin{spacing}{0.9}
    \item Issues: Non-mathematical topics.
        \vspace{-5pt}
        
Method: Use GPT-4o to judge, with the prompt outlined in the Appendix~\ref{sec:appendix:prompts}.
    \end{spacing}

\end{enumerate}

\end{spacing}
\vspace{-10pt}
\subsection{Standardization of Problems' Description}
\label{sec:appendix:B2}
\begin{figure}[h!]
  \centering
  \includegraphics[width=0.6\columnwidth]{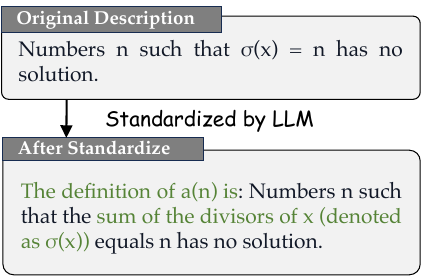}
  \caption{Comparison between original and standardized problem description. The standardized version includes  hints and explains the specific meaning of $\sigma(x)$.}
  \label{fig:standardized}
\end{figure}

\vspace{-10pt}
\subsection{Dataset Statistics}
\label{sec:appendix:B3}
\begin{figure}[h!]
    \centering
    \begin{subfigure}{0.4\linewidth}
        \centering
        \includegraphics[width=\linewidth]{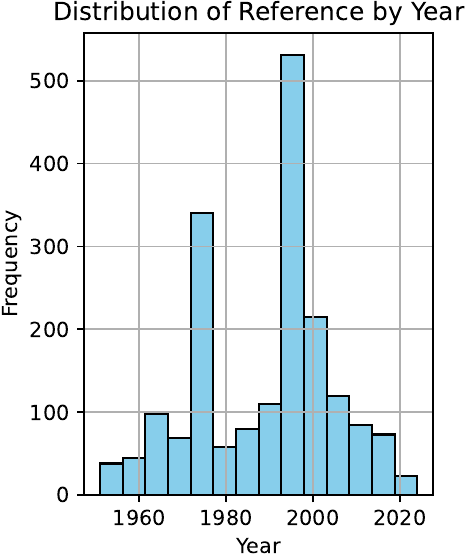}
    \end{subfigure}
    \hfill
    \begin{subfigure}{0.47\linewidth}
        \centering
        \includegraphics[width=\linewidth]{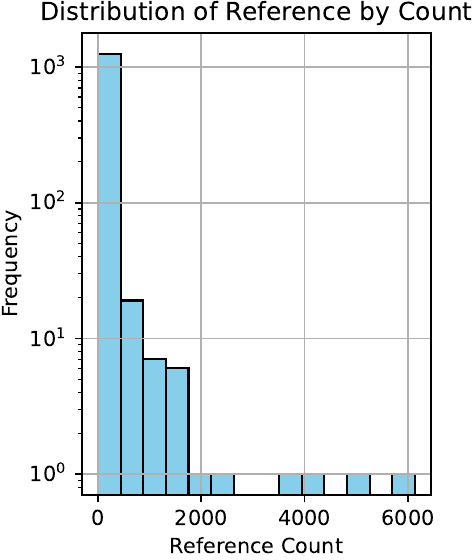}
    \end{subfigure}
    \caption{Distribution of references in UTMath.}
    \label{fig:distribution_utmath}
\end{figure}

To demonstrate that our benchmark is of cutting-edge level, we have analyzed the distribution of the publication years and the number of references included in the problems of the benchmark as shown in Fig.~\ref{fig:distribution_utmath}. Additionally, OEIS is a dynamic database. Over the past five years, more than 35,000 sequences in UTMath\_Train have been further researched, and over 2,000 new sequences have been added. This ongoing development makes it possible to continuously update UTMath\_Train and UTMath, helping to address the challenges posed by data leakage.

\newpage
\onecolumn 
\section{Prompts}

\label{sec:appendix:prompts}
\noindent
\begin{figure*}[h!]
    \centering
    \includegraphics[width=0.95\textwidth]{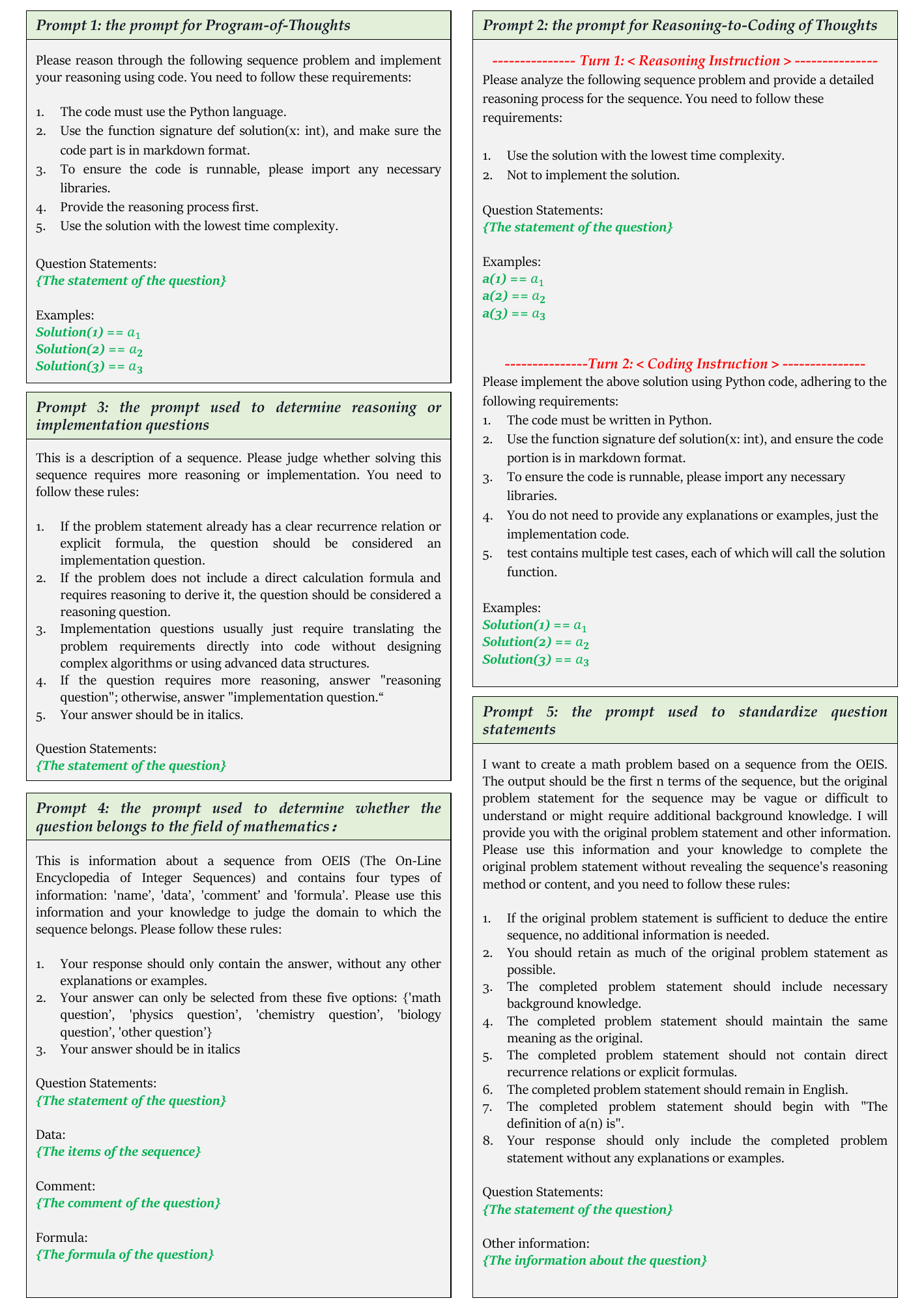}
    \captionof{figure}{The main prompts we used.}
\end{figure*}

\section{Case Studies}
\label{sec:appendix:Case Study}
\begin{figure*}[h!]
    \centering
    \includegraphics[width=0.78\textwidth]{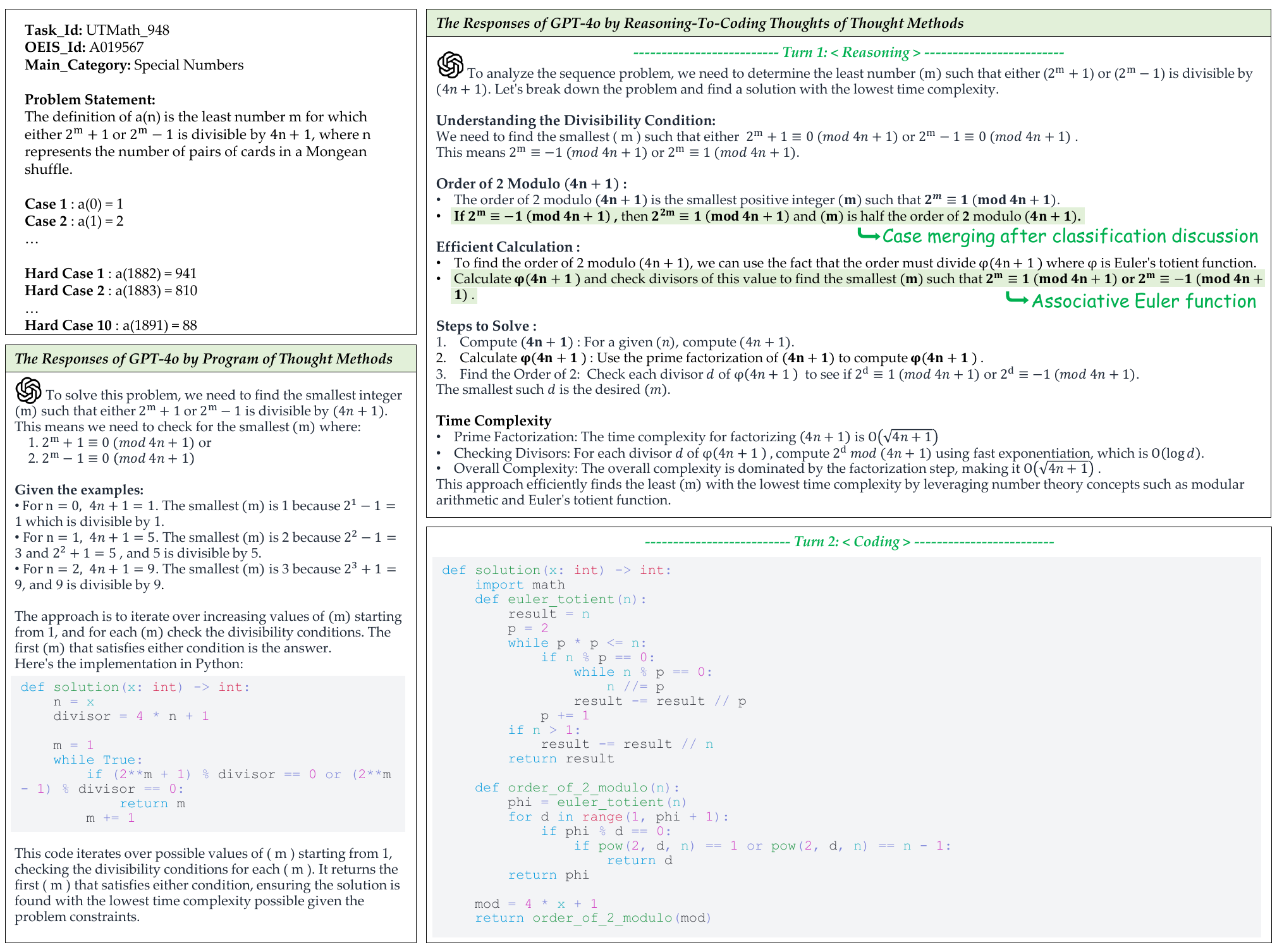}
    \caption{GPT-4o solves UTMath\_948 by the PoT method, by the RCoT method, respectively. The input prompt is omitted here but can be found in Appendix C. PoT simply performs brute-force solving, while RCoT involves deeper reasoning through Case merging after a classification discussion and the application of Euler's formula, providing a solution with lower time complexity.}
\end{figure*}
\begin{figure*}[h!]
    \centering
    \includegraphics[width=0.78\textwidth]{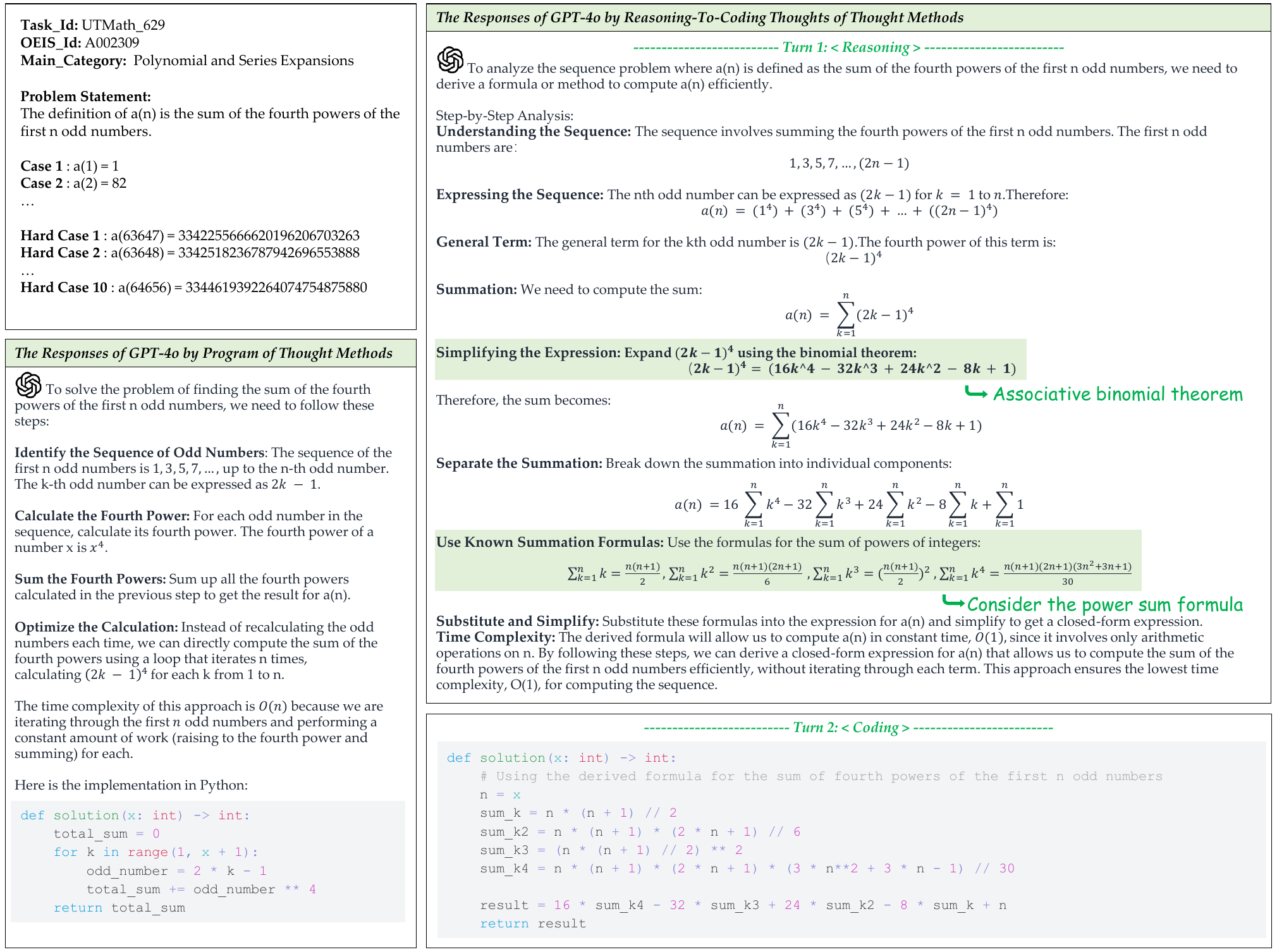}
    \captionof{figure}{GPT-4o solves UTMath\_629 by the PoT method, by the RCoT method, respectively. The input prompt is omitted here but can be found in Appendix C. PoT only performs brute-force traversal, whereas RCoT engages in deeper reasoning by associating the problem with the binomial theorem and using the power sum formula, yielding a closed-form expression with lower time complexity from O(n) to O(1).}
    
\end{figure*}

\end{document}